\documentclass[conference]{IEEEtran}
\IEEEoverridecommandlockouts
\usepackage{cite}
\usepackage{amsmath,amssymb,amsfonts}
\usepackage{algorithmic}
\usepackage{graphicx}
\usepackage[colorlinks=true,linkcolor=blue,citecolor=green]{hyperref}

\usepackage{textcomp}
\usepackage{xcolor}
\def\BibTeX{{\rm B\kern-.05em{\sc i\kern-.025em b}\kern-.08em
    T\kern-.1667em\lower.7ex\hbox{E}\kern-.125emX}}
\begin{document}

\title{The Rise of Creative Machines: Exploring the Impact of Generative AI\\

}

\author{\IEEEauthorblockN{Saad Shaikh}
\IEEEauthorblockA{\textit{dept. Btech in Information technology and Data Science} \\
\textit{Student at Ajeenkya D.Y Patil University}\\
Pune, INDIA \\
shaikh.saad@adypu.edu.in}
\and
\IEEEauthorblockN{Sakshi Mhaske}
\IEEEauthorblockA{\textit{dept. Btech in Information technology and Data Science} \\
\textit{Student at Ajeenkya D.Y Patil University}\\
Pune, INDIA \\
\vspace{10pt}
sakshi.mhaske@adypu.edu.in}
\and
\IEEEauthorblockN{Rajat Bendre}
\IEEEauthorblockA{\hspace{22pt}\textit{dept. Btech in Information technology and Data Science} \\
\textit{Student at Ajeenkya D.Y Patil University}\\
Pune, INDIA \\
rajat.bendre@adypu.edu.in}
\and
\IEEEauthorblockN{Dr. Ankita Aggarwal}
\IEEEauthorblockA{\textit{Corresponding Author} \\
\textit{Faculty at Ajeenkya D.Y Patil University}\\
Pune, INDIA \\
ankitahctm@gmail.com}

}

\maketitle
\begin{abstract}
Technology is being revolutionized by generative artificial intelligence (AI), which generates highly tailored and lifelike content automatically across a variety of media. Although this technology has the power to completely change businesses, it also poses social, legal, ethical, and security risks. This paper explores the practical uses of generative AI in research, product creation, and marketing by offering a thorough analysis of the field's state and future potential. It addresses important developments in the industry, such as the emergence of new players, the rapid expansion of text generation platforms, and the growing acceptability of creative generative AI. It also emphasizes the growing demand for user-friendly generative AI tools.

The paper discusses important moral concerns about misinformation, bias, employment displacement, and malevolent use of generative AI. It suggests mitigating measures like moral guidelines, legal frameworks, public awareness campaigns, and technological advancements and interventions unique to the industry. It offers a fair assessment, taking into account both the advantages and disadvantages of generative AI. The article's conclusion emphasizes the significance of responsible development, which calls for ongoing study and stakeholder dialogue to guarantee that generative AI has a beneficial social impact while reducing its negative effects.

\end{abstract}

\section{\textbf{Introduction}}
The development of generative AI signifies a radical change in the capabilities of technology. With little more than a text input, powerful machine learning models such as GPT-3 and DALL-E can now produce amazingly convincing text, graphics, music, video, and more. Future uses of this cutting-edge technology have the potential to transform a wide range of sectors, from expediting scientific research to building whole virtual worlds for entertainment. But generative AI also raises important issues that society is only now starting to address, including as security, intellectual property, ethics, and more. This study will examine the immense potential as well as the many problems that this quickly developing sector presents. The paper will look at significant advancements in the history of generative AI, ranging from early attempts at creating procedurally in video games to more current deep learning advances that allow for remarkably human-like creative output. Following this development, the paper will address cutting-edge businesses and research facilities that are presently setting new standards in this field, as well as technologies that have enhanced capabilities like text and picture synthesis. Following an overview of the history and state-of-the-art methods in generative AI, the paper will present specific instances of its application in a variety of industries, including the creative arts, healthcare, and education. Generic artificial intelligence (AI) has the potential to transform various industries, such as drug discovery and tutoring, provided it is carefully incorporated and controlled. The risks posed by the widespread use of these systems, including algorithmic bias, disinformation, and intellectual property infringement, will also be critically examined in this paper. The investigation of the pressing issues regarding ethics and security raised by generative AI will be informed by recent controversies surrounding defective outputs and the risks of widespread media manipulation. The overall goal of this paper is to provide a thorough overview of this quickly developing technology's past, present, and potential future. We can try to improve generative AI's advantages while proactively reducing its risks by carefully examining its potential and highlighting areas that call for caution. This thoughtful examination of the great potential as well as the many hazards will add depth to important conversations about getting ready for the responsible development and application of one of the most revolutionary technologies to emerge in decades.

\section{\textbf{Extent and Impact of Generative AI}}

\subsection{\textbf{AI Transforming Industries and Consumers}}
\begin{figure*}
    \centering
    \includegraphics[width=0.7\linewidth]{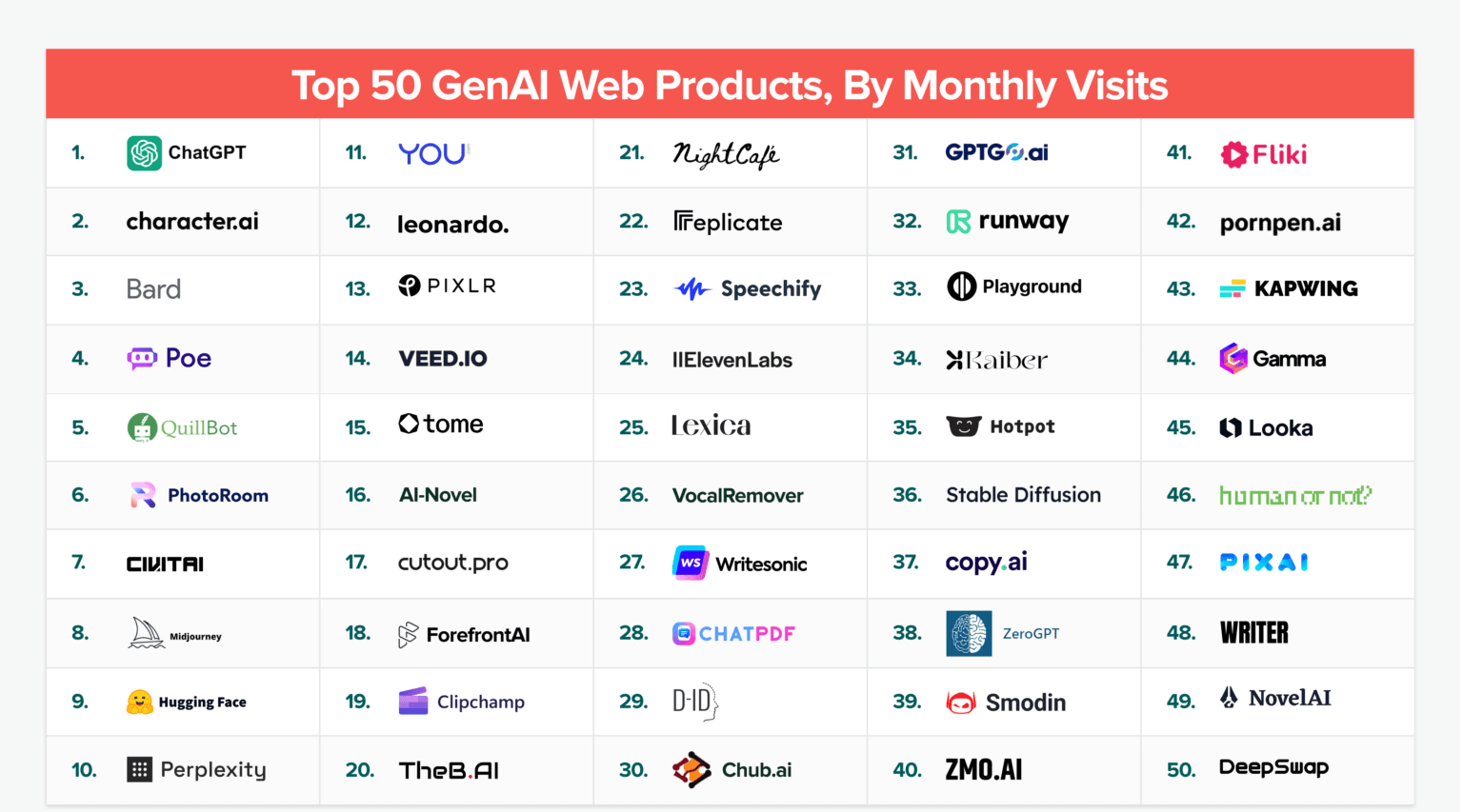}
    \caption{\textbf{Generative AI Products}}
    \href{https://a16z.com/how-are-consumers-using-generative-ai/}{a16z.com}
    \label{1}
\end{figure*}

\begin{figure*}
    \centering
    \includegraphics[width=0.7\linewidth]{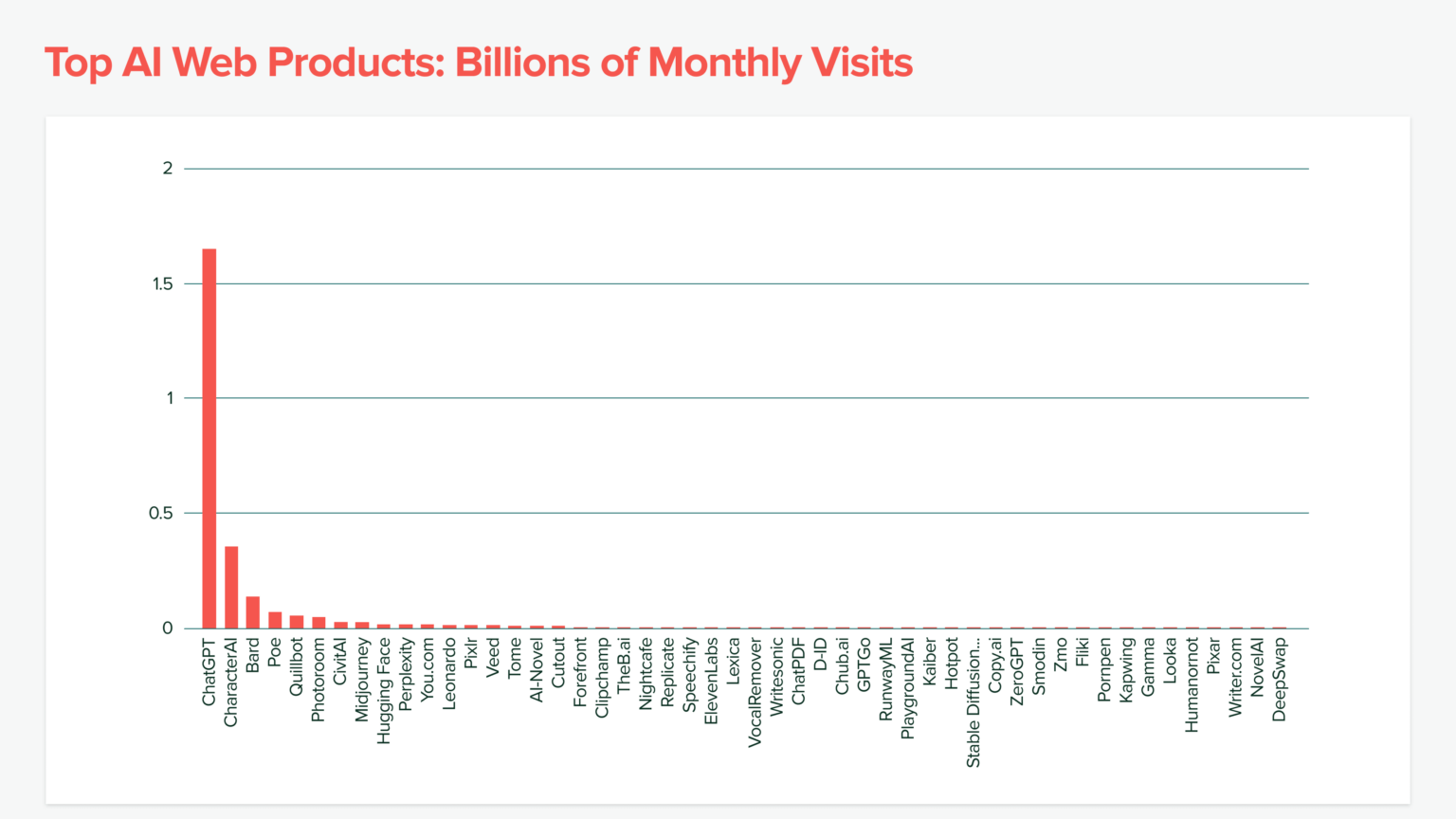}
    \caption{\textbf{Growth Of Different AI Products}}
    \href{https://a16z.com/how-are-consumers-using-generative-ai/}{a16z.com}
    \label{2}
\end{figure*}

\begin{figure*}
    \centering
    \includegraphics[width=0.7\linewidth]{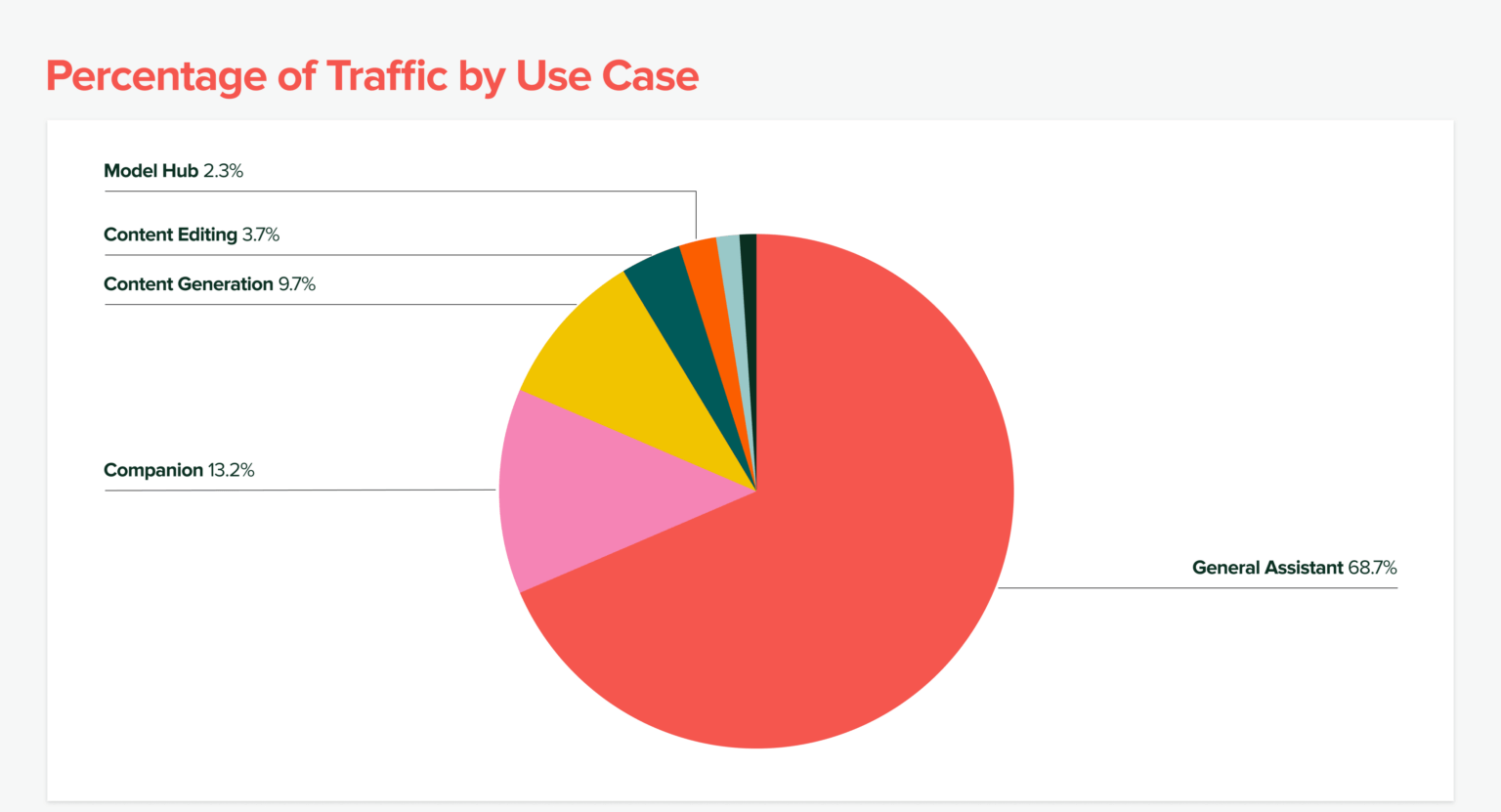}
    \caption{\textbf{Popularity Via Usage}}
    \href{https://a16z.com/how-are-consumers-using-generative-ai/}{a16z.com}
    \label{3}
\end{figure*}

Generative AI is already being used in a variety of real-world applications that are transforming industries and impacting consumers. Some key areas where generative AI is being applied includes the products shown in the ``Fig.~\ref{1}''.

Using generative AI, text, photos, audio, and video content can all be produced automatically. Businesses use this capability to produce music, art, blog posts, articles, social media posts, and marketing materials. The potential for generative AI to enable rapid, personalised, and low-cost content creation could upend entire industries, such as graphic design, music composition, and journalism. Product design and development can be completed more quickly and effectively thanks to generative AI. It is employed in the design of new consumer goods, including furniture, clothes, medications, and other items. Businesses can accelerate the release of innovative products onto the market by automating certain aspects of the design process. Workflows and economics in product development could be greatly impacted by this technology. Scientists benefit from generative AI's ability to suggest novel theories and research directions. It has already been applied to the creation of fresh drug candidates and substances. Generative AI may quicken the pace of innovation across all scientific fields by supporting human researchers. These generative AI capabilities are being leveraged by both startups and major corporations. Prominent instances comprise OpenAI's DALL-E 2 for producing lifelike images from text, Anthropic's Claude for engaging in natural language conversations, and Alphabet's AlphaFold for protein structure prediction. Rapid consumer adoption of generative AI is also being made possible by the introduction of tools like OpenAI's GPT-3 language model.

An examination of the top 50 consumer-facing generative AI web products reveals that: - Text, image, and code generation are the most widely used features.

- The generation of text and images is led by incumbents.

- Code generation is becoming more popular among new startups.

- Over 200 percent more traffic has been coming to these products on average over the last 12 months.
According to this analysis, there is a sizable demand from users for intuitive generative AI applications. The next "big winner" in the generative AI market might be the business that can effectively blend functionality and power in a single product.

\section{\textbf{Insights from Top Generative AI Companies}}
Rapid advancements in generative AI have occurred in recent years, leading to the emergence of new products and capabilities that are changing markets and affecting consumers. Several significant trends are revealed by analysing the top 50 consumer-facing generative AI companies:Eighty percent of the top 50 companies are new, having launched within the last year. This demonstrates how quickly generative AI is developing. 48 percent of the companies are fully self-funded startups, and only five have ties to large tech companies. This raises the prospect of developing innovative AI products with little outside funding.  

\textbf{ChatGPT Has Early Dominance}
Currently, ChatGPT completely rules the consumer AI market. It is the 24th most visited website worldwide as of June 2023, receiving 1.6 billion visits each month. CharacterAI, the second-biggest player, only accounts for 21 percent of ChatGPT's traffic. ChatGPT is still smaller than most mainstream websites, comparable to sites like Reddit and LinkedIn. However, it has grown extremely quickly.checkout the ``Fig.~\ref{2}''.

\textbf{Text Generation Leads, Creative Tools Rising }
Chatbots like ChatGPT account for 68 percent of consumer traffic, making text generation the most popular application. However, creative tools like image, music, and video generation are rapidly gaining traction. Image generation has 41 percent of creative tool traffic, writing tools have 26 percent, and video generation 8 percent as shown in the ``Fig.~\ref{3}''.

\section{\textbf{Risks of Generative AI}}
\subsection{\textbf{Balancing Progress and Responsibility}}
Though a revolutionary technology, generative AI is not without risk. The possibility of false information and deepfakes being created is one major worry. Images, videos, and audio recordings can all be produced by generative AI that will look very real to the viewer. Misinformation and malicious impersonation could be disseminated through this. The automation of tasks currently performed by human professionals like journalists, copywriters, and graphic designers poses another risk of job displacement. This is due to the development of generative AI.

\begin{figure*}
    \centering
    \includegraphics[width=0.7\linewidth]{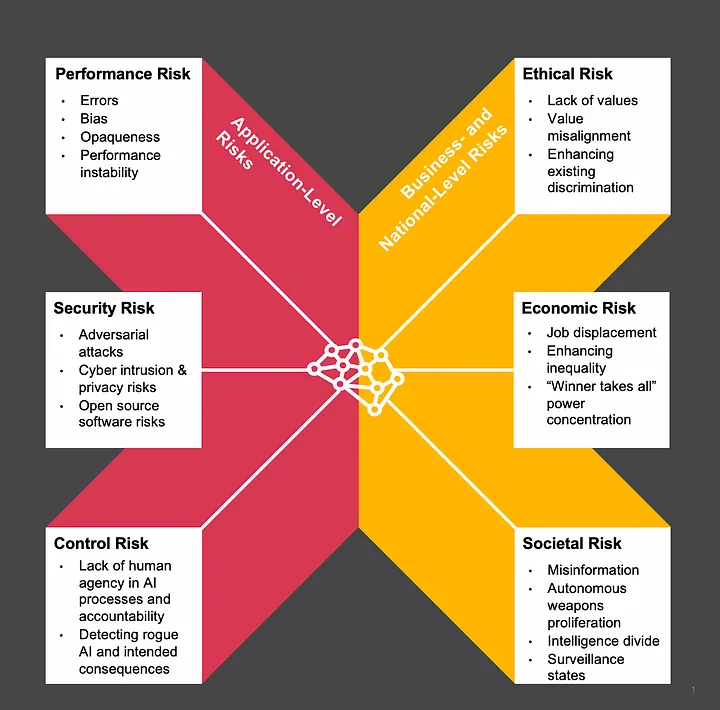}
    \caption{\textbf{Risks Of Generative AI }
    }
     \href{https://towardsdatascience.com/five-views-of-ai-risk-eddb2fcea3c2}{towardsdatascience.com}

    \label{4}
\end{figure*}

Furthermore, prejudice and discrimination can still affect generative AI systems. Given that they receive training on large-scale datasets, any biases inherent in the data may find their way into the content they produce, which may serve to reinforce negative stereotypes. Developing techniques to recognise and stop the spread of false content is essential to mitigating these risks. Furthermore, it is crucial to guarantee that generative AI systems are trained on objective data and to encourage their ethical and responsible use. Although generative AI has the potential to revolutionise a number of industries and aspects of our lives, we must constantly be aware of the risks it entails and take proactive steps to address and mitigate them.
With its rapid advancement and potential to transform industries and societal aspects, generative AI demands a thorough investigation of its causes and effects in order to ensure responsible and ethical application. The proliferation of large datasets, which enable models to understand complex patterns and relationships, is one of the main factors contributing to the rise of generative AI. The development of reliable computing hardware, which enables the computational demands of training and using these models, is equally important. This rise is further aided by the development of machine learning algorithms, the open-source movement that facilitates AI development, and the widespread use of cloud computing.
Significant effects of generative AI are already being seen in a variety of industries. Creative domains give rise to avant-garde literature, music, and art because they produce realistic images, melodies, and literary works. Within the business domain, it streamlines processes through the automation of tasks, improvement of efficiency, and provision of insightful data. Customer service, product innovation, and marketing content are the main winners. Healthcare, meanwhile, is using generative AI to diagnose diseases, develop new drugs, and customise treatments. This approach promises individualised treatment regimens and inventive targets for new drugs.Benefits of generative AI include increased productivity and creativity, easier problem solving, and better quality of life. Contrarily, its more sinister potentialities include the spread of false information and disinformation, raising issues with democracy and public confidence. Due to task automation, job displacement is a significant concern. Another unsettling possibility is the production of dangerous or harmful content, such as deepfakes and autonomous weapons. In summary, the complex causes and consequences of generative AI highlight its revolutionary potential, necessitating close supervision to maximise its advantages and minimise its hazards.

In conclusion, generative AI's complex causes and effects highlight both its potential for positive change and its potential to pose significant societal challenges. Enabling its advantages while reducing its risks depends critically on its responsible development and application.

\subsection{\textbf{Mitigating Risks and Facing Oppositions}}

When discussing the dangers of generative AI, a number of viable remedies and counterarguments surface. A multifaceted strategy is required to effectively mitigate these risks. First and foremost, it is crucial to establish and follow ethical standards for the development and use of generative artificial intelligence. To ensure responsible AI development, these guidelines should address issues of bias, privacy, security, and transparency.
Furthermore, regulation is important. In delicate fields like politics, where deepfakes could be extremely dangerous, governments can enact regulations to regulate the application of generative AI. Additionally vital is public education. People can make wise decisions, spot false information, and effectively combat disinformation by developing a better understanding of generative AI.
In addition, there is a keen pursuit of technical solutions. Researchers are working to create watermarks to distinguish genuine content from deepfakes as well as novel techniques for identifying and preventing fake news and disinformation.
In addition to these general tactics, solutions specific to the industry must be taken into account. For instance, generative AI has the potential to completely transform the development of individualised treatment plans in the healthcare industry. However, in order to prevent bias, models must be trained on high-quality, diverse, and representative data. Similar to this, generative AI can be an effective tool in the financial industry for identifying fraud and money laundering, which calls for the installation of strong security measures to guard against abuse.
It's important not to undervalue how ethical generative AI is. It is imperative to take into account factors like prejudice, privacy, security, and openness. Preventing discrimination and negative consequences requires addressing bias in training data. When realistic images and videos are created, privacy concerns surface, and precautionary steps like gaining consent are required. Transparency is key to building trust, with complete disclosure of data sources, algorithms, and potential hazards. Security measures are necessary to prevent the production of malicious content, such as deepfakes.
A spectrum of opposing perspectives on generative AI exists amid these factors. It is argued by some that this technology is dangerous because it can produce hate speech, fake news, and autonomous weapons. It can also threaten jobs by automating jobs. Another issue to be concerned about is the possibility of bias in AI models.
On the other hand, generative AI advocates view it as a potent instrument that can boost productivity and creativity by automating processes, resolving challenging issues, and enhancing general quality of life. It is capable of coming up with fresh concepts, producing imaginative writing, and providing original answers to pressing societal issues like creating novel medications and therapies, planning effective transit networks, and completely revamping the educational system. In conclusion, industry-specific precautions, laws, regulations, technical advancements, and ethical standards are all necessary to reduce the risks associated with generative AI. Though there are legitimate worries about its abuse and bias, generative AI has enormous potential to improve society overall by increasing creativity and problem-solving skills as well as general quality of life. To fully utilise this revolutionary technology, it will be necessary to strike a balance between these divergent viewpoints.

\section{\textbf{Additional thoughts}}
\subsection{\textbf{Personal Observations}}
The world is embracing AI and seeing its potential in what it can accomplish thanks to generative AI, which is revolutionising the field. For instance:

1. Can be used to build lifelike digital twins of individuals that can be used for training in a variety of fields, including education, healthcare, and customer service.

2. Make artificial data that can be used to train other AI models and enhance their performance.

3. Customise learning experiences for each student by creating activities and content that are specific to their needs. This is how the renowned Khan Academy achieved it by using GPT 4 behind the scenes to assign each student to a personal tutor, which lowers costs and improves student learning.
Noting that generative AI is still a relatively new technology and that its full potential is still unrealized is important. We can anticipate much greater effects on society as generative AI develops and becomes more advanced. which may have a significant impact on how society functions.

\subsection{\textbf{Call To Action}}
With the potential to impact every aspect of our lives, from the trivial to the significant, generative AI has the power to drastically alter our world.A thorough and ongoing conversation must be had about its implications as its capabilities continue to advance at an unprecedented rate.
This study has acknowledged the risks and difficulties associated with generative AI while also highlighting its enormous potential. Still, we only have a limited grasp of this emerging technology. Much more study is required to fully understand its implications and make sure it is used for the good of humanity.
For this reason, we are putting out a call to action to researchers, decision-makers in government, business executives, and the general public to join us in this vital project. In particular, we implore:

• Scholars investigate the advantages, disadvantages, and biases of generative AI by delving deeper into its technical foundations. Additionally, we must work to create strategies for guaranteeing the explainability, safety, and security of generative AI systems.

• Legislators should think about the societal, legal, and ethical ramifications of generative AI. Creating frameworks to handle matters like liability, ownership, and intellectual property is part of this.

• Responsible practises in the development and application of generative AI should be adopted by industry leaders. This entails speaking openly with stakeholders about the application of generative AI and addressing any concerns they may have.

• The general public should learn about generative AI and take part in conversations about its future. This entails exercising critical thought when utilising generative AI and being aware of its possible advantages and disadvantages.

Together, we can make sure that generative AI is put to good use and contributes to the development of a more fair, just, and prosperous future for all.
Let's not watch as this technological revolution unfolds as spectators. Let's design a future in which generative AI drives progress and positive change.

\section{\textbf{Conclusion}}
Generative AI is poised to redefine creativity, reshape industries, and change the very foundation of society as we know it. It stands at the cusp of a transformative era. It has broad and far-reaching implications that include both tremendous potential and significant challenges.
Generative AI has the potential to unleash previously unheard-of levels of creativity and productivity. It can free up human ingenuity for more creative and strategic endeavours by automating tasks and producing new ideas. In disciplines like drug discovery, materials science, and design, generative AI can quicken the pace of research and result in discoveries that would be unthinkable otherwise. Furthermore, generative AI holds promise for democratising creativity and opening it up to a larger audience. Generative AI can enable people to express themselves in captivating new ways by offering tools to help with ideation, execution, and refinement. This might result in an explosion of artistic expression and a more inclusive and diverse creative community.
But the emergence of generative AI also brings up a number of issues. The possibility of losing one's job is among the most urgent. Work that is currently done by humans will probably be automated by generative AI as it develops. Social unrest and widespread unemployment could result from this. The potential for abuse is another issue. Disinformation such as deepfakes and fake news can be produced using generative AI. This could sow discord in society and erode trust in institutions. It is crucial to develop ethical standards for the creation and application of generative AI in order to reduce these risks. These rules ought to cover things like accountability, transparency, and bias. In the end, generative AI has significant and far-reaching effects. It could be advantageous to society or detrimental. The secret is to maximise its positive effects while reducing its negative ones.

Generative AI is a two-edged sword, to sum up. While it also presents risks to jobs and social stability, it has the potential to bring about a more creative and prosperous world. Our use of generative AI will determine how it develops in the future.


\begin{thebibliography}{00}


\bibitem{b1}Chris Stokel-Walker’s and Richard Van Noorden’s,"The Promise And The eril of Generative AI", Springer Nature Limited, Vol.614, 9 Feb 2023.
\bibitem{b2}David Baidoo-Anu and Leticia Owusu Ansah, "Education in the Era of Generative Artificial Intelligence (AI): Understanding the Potential Benefits of ChatGPT in Promoting Teaching and Learning," Journal of AI, vol. 7, no. 1, January-December 2023.

\bibitem{b3} Weng Marc Lim, Asanka Gunasekara, Jessica Leigh Pallant, Jason Ian Pallant, Ekaterina Pechenkina, "Generative AI and the Future of Education: Ragnarök or Reformation? A Paradoxical Perspective from Management Educators," The International Journal of Management Education, vol. 21, no. 2, July 2023, p. 100790.

\bibitem{b4}Yogesh K. Dwivedi, Nir Kshetri, Laurie Hughes, Emma Louise Slade, Anand Jeyaraj, Arpan Kumar Kar, Abdullah M. Baabdullah, Alex Koohang, Vishnupriya Raghavan, Manju Ahuja, Hanaa Albanna, Mousa Ahmad Albashrawi, Adil S. Al-Busaidi, Janarthanan Balakrishnan, Yves Barlette, Sriparna Basu, Indranil Bose, Laurence Brooks, Dimitrios Buhalis, Lemuria Carter, Ryan Wright, "So what if ChatGPT wrote it? Multidisciplinary perspectives on opportunities, challenges and implications of generative conversational AI for research, practice, and policy," International Journal of Information Management, vol. 71, August 2023, p. 102642.
\bibitem{b5} Mladan Jovanovic and Mark Campbell, "Generative Artificial Intelligence: Trends and Prospects," IEEE, October 2022.

\bibitem{b6} Jonas Oppenlaender, Aku Visuri, Ville Paananen, Rhema Linder, Johanna Silvennoinen, "Text-to-Image Generation: Perceptions and Realities," arXiv:2303.13530v2 [cs.HC], 2 May 2023.

\bibitem{b7}Chen Chen, Jie Fu, Lingjuan Lyu, "A Pathway Towards Responsible AI Generated Content," arXiv:2303.01325v2 [cs.AI], 17 Mar 2023.

\end{thebibliography}
\end{document}